\pdfoutput=1
\documentclass[11pt]{article}

\usepackage[final]{acl}

\usepackage{times}
\usepackage{latexsym}
\usepackage{adjustbox}

\usepackage[T1]{fontenc}
\usepackage[utf8]{inputenc}

\usepackage{microtype}

\usepackage{inconsolata}

\usepackage{graphicx}

\usepackage{todonotes}           % use this line to show todos

\title{Recent Trends in Linear Text Segmentation: A Survey}

\author{Iacopo Ghinassi\\
  \small{Queen Mary University of London / London, UK} \\
  \small{\texttt{i.ghinassi@qmul.ac.uk}} \\\And
  Lin Wang \\
  \small{Queen Mary University of London / London, UK}\\
  \small{\texttt{lin.wang@qmul.ac.uk}} \\\AND
  Chris Newell\\
  \small{BBC R\&D / London, UK} \\
  \small{\texttt{chris.newell@bbc.co.uk}}\\\And
  Matthew Purver \\
  \small{Queen Mary University of London / London, UK} \\
  \small{Institut Jožef Stefan / Ljubljana, Slovenia} \\
  \small{\texttt{m.purver@qmul.ac.uk}}}

\begin{document}
\maketitle
\begin{abstract}
    Linear Text Segmentation %aims at
    is the task of
    automatically tagging text documents with topic shifts, i.e.\ the places in the text where the topics change. %The field of Linear Text Segmentation 
    A well-established area of research in Natural Language Processing, drawing from well-understood concepts in linguistic and computational linguistic research, the field has recently seen a lot of interest as a result of the surge of text, video, and audio available on the web, which in turn require ways of summarising and categorizing the mole of content for which linear text segmentation is a fundamental step. In this survey, we provide an extensive overview of current advances in linear text segmentation, describing the state of the art in terms of resources and approaches for the task. Finally, we highlight the limitations of available resources and of the task itself, while indicating ways forward based on the most recent literature and under-explored research directions.\footnote{Chris Newell is no longer part of BBC R\&D, but the paper was authored when still working at the organisation.}
\end{abstract}

\section{Introduction}

Linear text segmentation, also known as topic segmentation, is the task of identifying topic boundaries in a text using coherence modeling and/or local cues \cite{Purver2011}. The attribute `linear' derives from the fact that in this setting, which is the most popular but not the only one, topics are considered ``linearly'' as following one another in documents and, as such, \textit{linear} text segmentation ignores any sub-topic or hierarchic structure and focus on finding the boundaries between the topics thus linearly defined. This is also distinguished from topic \textit{classification}, which relates to classifying text with the correct topic class; while linear text \textit{segmentation} is strictly tasked with identifying the part of a text in which a topic boundary occurs. Such boundaries then have a relevant role in a variety of contexts, such as finding individual news stories in a news show or podcast \cite{Ghinassi2021} or even as a pre-processing step for tasks like summarization \cite{zhong-etal-2021-qmsum}. 

This survey aims to give a comprehensive, yet brief overview of the field, highlighting the evolution of the approaches used to tackle the task as well as the available metric and resources and what remains to be done. Such a survey is much needed as previous surveys on the topic are mostly outdated at this point \citep[see, e.g.,][]{Purver2011}. 
% while the survey by \cite{Pak2018} does not even focus solely on topic segmentation, but rather gives a broader picture of different types of text segmentation (e.g.\ word, sentence, phrase topic segmentation, etc.). 
Crucially, previous surveys lack an in-depth exploration of the use of language models for the task, where transformer-based language models and Large Language Models (LLMs) have now become, as in other areas of NLP, central for the task. In this survey, then, we aim to fill this gap by showing how the field has slowly shifted to use features from transformer-based language models and supervised learning as the framework of choice and how LLMs are just starting to get traction. In doing so, we will also highlight the various problems of resources and evaluation which, we argue, are central for further developments in the field. Finally, we discuss future directions. 

This work is a necessary step for summarising and grounding recent research in the field, while pointing towards future developments which are worth the focus of future research. %At the same time, 
Note that this survey does not touch upon sub-areas like multi-modality and more niche domains like video-lecture segmentation: we focus on NLP and on the domains in which topic segmentation has traditionally been seen as a central task. Future research might integrate the current work with these aspects. % that were left off.

\section{Linear Text Segmentation Approaches}
% \subsection{Rule-based Systems}
% Systems based on hand-written rule specific, for example, to the format of a news story \cite{Wang2003, Chua2004}. If a news show presents a topic caption that changes every time a news story is introduced, for example, the simplest way to identify a topic shift in this scenario is to use an optical character recognition system over the portion of the image in which the caption appears\cite{Wang2003}. Such methods, however, are totally dependent on the specific format of the input programme and they provide no way of learning a deeper semnatic understanding of the content itself. This is why this work focuses on statistical, machine learning techniques to the problem of topic segmentation. 

\subsection{Basic Units}
A first step in designing a linear text segmentation system is deciding which basic unit of text to use as input to the system. Generally, linear text segmentation systems work either at the word, sentence (or pseudo-sentence), or paragraph level.

Research in discourse structure has highlighted that paragraphs usually play crucial roles in conveying different topics in written text \cite{Cohesion1976,DiscourseStructure} and, as such, early literature often used %such a 
the paragraph as the unit \cite{HierarchicalTopicSeg}. As the technology started being applied to domains such as multimedia content, spoken language, and, in general, text not having paragraph information, however, the role of paragraphs as preferred basic units was progressively superseded by textual features corresponding to words and sentences; or, in early literature, \textit{pseudo-sentences}, %in early literature, where
in which 
an arbitrary number of words are aggregated to avoid introducing error from sentence tokenization (%which is 
now largely a solved task for languages such as English). In the case of multi-speaker scenarios such as most meeting transcripts the preferred basic units are usually %consists of 
\textit{speaker turns}, as segments that are usually sufficiently complete to represent coherent units or at least to convey the communicative intention shared by speaker and hearer, but systems working at the word level have been widely used as well. 

Currently, the preference for using word or sentence-based methods seems to be mostly dependent on the type of features being used in end-to-end systems. Models built on word-topic probability distributions \cite{Purver2006,Sun2008,Misra2011} or word embeddings, then, use words as basic units \cite{Koshorek2018,arnold-etal-2019-sector,yu-etal-2023-improving-long}, while models built on sentence embeddings employ sentences or speaker turns \cite{Ghinassi2021,Ghinassi2023,FacebookArticle}.

\subsection{Unsupervised Methods}
\subsubsection{Count-based Methods}
One of the earliest %among 
unsupervised techniques for linear text segmentation, TextTiling, used two adjacent sliding windows over sentences and compared the two blocks of sentences inside these windows using cosine similarity between the relative bag-of-words vector representations \cite{Hearst1994}. The same algorithm has been successfully used with different, more informative sentence representations, such as TF-IDF re-scoring of bag-of-words  \cite{Galley2003}.
% and features derived from generative topic models like Latent Dirichlet Allocation (LDA) \cite{Riedl2012}. The TextTiling algorithm is still used in the present days as a baseline system to try new, potentially better sentence representations from different modalities like neural sentence encoders \cite{Ghinassi2021, Harrando2021}.
% with later literature substituting them with  representations from generative topic models \cite{Misra2011, Sun2008}. 
To further improve the individuation of topically incohesive adjacent windows of sentences, the C99 algorithm was proposed \cite{choi2000}. This method builds on the intuitions of TextTiling but substitutes the step in which the similarities are scored with a divisive clustering algorithm, improving over the original approach. 
% Using matrix factorization over the bag-of-words sentence representations (i.e.\ Latent Semantic Analysis) to derive the features to be used in C99 improved results even more \cite{choi2000}, further proving the importance of meaningful, more abstract sentence representations in topic segmentation systems.

Another early approach in topic segmentation was that of using the distance between sentence representations in a dynamic programming framework, including Hidden Markov Models (HMMs). Count-based language models (i.e.\ n-gram models) were proposed in this context, where the probability of different words under different topics has been used either directly in an HMM framework \cite{EarlyHMMTopSeg} or using a linear dynamic programming approach as in the U00 system \cite{Utiyama2001}. The most recent approach in this sense, BayesSeg, added probabilistic models of cue phrases to a count-based language model, reaching results that are still competitive \cite{Barzilay2008}. The use of language models, even though in a radically different way, is at the base of the most recent segmentation systems. 

\subsubsection{Topic Modelling Methods}
Early on, researchers combined techniques from the closely related task of topic modelling to perform topic segmentation. The use of topic models for the task falls broadly into the category of generative topic segmentation models, as it shifts the focus from discriminatively identifying areas of low cohesion and local cues, to directly modeling the underlying topics ``generating'' the different segments in the document \cite{Purver2011}. 

Most early approaches in this sense build on various forms of Latent Dirichlet Allocation (LDA) as a method to automatically individuate topics in text via count-based features \cite{Blei2003}. LDA produces, among its outputs, a matrix of word-topic assignments, storing the probability of each word in the given vocabulary under different topics.
Dynamic programming approaches have been widely used in this context. The MM system, for example, used such a framework in conjunction with probabilities derived from word-topic assignments to decide over the most likely topic at each word in the sequence \cite{Misra2011}. 

More recently, TopicTiling used word-topic assignments from LDA models to create word vectors and, by aggregating word vectors, sentence vectors to be used as sentence representations for the TextTiling algorithm \cite{Riedl2012}.

An advantage of using topic modelling as a base for topic segmentation is that such algorithms automatically yield the classification of topic segments as a by-product, as the probability associated with different topics can be aggregated at the segment level after segmentation \cite{Purver2006}. Using generative topic models also makes it easier to tackle the task in a hierarchical fashion, where the level of granularity of the topics (and therefore of the segmentation) can be directly controlled \cite{du-etal-2013-topic}. These are indeed properties that do not yet have a parallel in modern end-to-end systems and, as we will see, combining the two paradigms is a research direction worth pursuing.

\subsubsection{Embeddings-based Methods}

Another more recent strand of research has drawn from improvements in vector semantics and initially used word embeddings to determine the coherence of consecutive words in the context of topic segmentation. This concept has been variously applied 
% either in combination with dynamic programming \cite{Alemi2015} or 
in algorithms such as GraphSeg \cite{Glavas2016}, comparing consecutive sentences based on a graph of similarities between their constituent word embeddings.

More recently, the evolution of neural language models has shifted the paradigm from word-based methods to sentence-based ones, in which dense sentence representations are obtained from transformer-based language models like BERT and employed in conventional techniques such as TextTiling \cite{Ghinassi2021,FacebookArticle}.

% Generally, embedding-based methods have not shown massive improvements over more traditional baselines, especially in the context of segmentation of dialogue, as evident from table \ref{tab:icsiresults} discussed below in more details.
% Similarly drawing from LLMs, transfer learning have also been applied by using the next sentence prediction capabilities innate in BERT to obtain coherence scores of consecutive utterances in meeting transcripts, which can then be used instead of cosine similarities in the TextTiling framework \cite{xing-carenini-2021-improving}.

% In general the use of LLMs to extract features have also allowed for smaller datasets necessary to train supervised systems which have largely outperformed unsupervised systems in most domains of application.

% The use of graphs in the context of topic segmentation is itself an active area of research with early literature proposing the use of graph algorithm from computer vision such as normalized cuts \cite{Shi2000}. Normalized cuts have been used in topic segmentation with lexical features \cite{Malioutov2006}.

\subsubsection{LLM-based Methods}

During last year, pioneering work has also been carried out using multi-billion parameter LLMs such as ChatGPT and prompt engineering to treat the problem as a Natural Language Generation (NLG) task \cite{ChatGPTTopSeg,yu-etal-2023-improving-long}. The use of LLMs in a zero-shot setting can be %inscribed in the unsupervised methods 
classed as an unsupervised method, and it has been shown to outperform all other unsupervised methods after careful prompt optimization \cite{ChatGPTTopSeg,jiang-etal-2023-superdialseg}. This approach, then, is promising and it should be explored as a way forward to overcome specific limitations of the generally more effective supervised framework described below.

\subsection{Supervised Methods}
Supervised methods have been present since early on in the field. The surge of these methods, however, coincides with the improvements in neural language modeling and, as such, we limit our description to such methods. For an in-depth discussion of discriminative supervised methods before neural language models, we refer to \cite{Purver2011}.

\subsubsection{Single-Task Methods}
As mentioned, advances in neural language models have changed also the landscape of linear text segmentation, as they did for NLP more generally. In the context of linear text segmentation, this meant a progressive shift towards supervised end-to-end systems (typically based on neural architectures) building on strong semantic features like modern word and sentence embeddings, as well as new large datasets to train such systems.

In the supervised setting, the segmentation problem is often treated as one of sequence tagging, where a binary scheme is used to label individual units such as sentences, to individuate where a segment ends or starts.

Among the first such approaches, TextSeg \cite{Koshorek2018} is a hierarchical LSTM model that builds on Word2Vec features and that outperformed by a large margin other methods available at the time. Following this work, other systems have been proposed similarly building on recurrent neural networks and word embeddings, with several improvements either at the embedding level \cite{arnold-etal-2019-sector} and/or at the classifier level \cite{Badjatiya2018,Sehikh2018}.

As transformer-based language models changed the landscape of NLP, transformer and LSTM classifiers for linear text segmentation drawing on sentence-level BERT features started being proposed as well \cite{Lukasik2020,xing-etal-2020-improving} and they have since become the norm, as they have been shown to outperform other features for the task \cite{GhinassiPeerJ}. The use of pre-trained language models like BERT to extract features (generally known as transfer learning) has been shown to improve the generalization capabilities of topic segmentation systems, thanks to the general knowledge encapsulated in such encoders. 

LSTM architectures building on such features have been shown to outperform Transformers for the task in certain cases, especially when not enough training data is available \cite{Ghinassi2023}, while they perform comparatively similarly in case of bigger datasets \cite{Lukasik2020}. This evidence also reflects the tendency of such models to overfit to specific cue phrases and domain-specific features  \citep[e.g.\ naming a correspondent in certain news shows,][]{GhinassiPeerJ} and the use of domain adaptation has also been proposed in this context to attenuate the problem of overfitting to specific domains that come with the supervised setting \cite{glavas-etal-2021-training}. 

Finally, a very recent line of research has attempted to use transformer-based language models directly as classifiers by placing a linear classification head on top of the beginning of sentence tokens. Among the limitations of transformers is the quadratic cost of self-attention that severely limits the maximum input length in terms of tokens for models like BERT. Earlier systems like Cross-segment BERT initially limited the context available to BERT by inputting just pairs of sentences \cite{Lukasik2020} or passing sliding windows over tokens to aggregate as much context as possible \cite{SlidingWindowBERT}. More recent works have used models such as Longformer, specifically designed to deal with long contexts to overcome this problem \cite{Inan2022StructuredSU}. 
% The same work also proposed to frame the problem as a text-to-text generation problem, where the encoder-decoder paradigm can be used to generate the position of the topic boundaries directly as tokens by a transformer-based decoder (e.g.\ 2 if the second sentence corresponds to a topic boundary).

\subsubsection{Multi-task Methods}
A more recent trend in linear text segmentation systems has variously adopted multi-task learning to regularise and improve end-to-end systems. Among the drawbacks of existing end-to-end systems, it has been observed how such models tend to overfit on local, domain-dependent cues that signal topic shifts (e.g.\ the locution ``moving on'' in multi-party meetings), but often do not generalize to other domains \cite{GhinassiPeerJ}. In this sense, multi-task learning works similarly to transfer learning in helping the model to extract more general features, which more closely relate to modeling the underlying topical coherence.

Systems belonging to this category mostly combine topic classification and topic segmentation, both framed as supervised tasks. Topic classification in this context is framed as the task of assigning the correct topic class to each sentence or basic unit in the text, rather than identifying the basic units which are topic boundaries (i.e.\ linear text segmentation). This strand of research emerged mostly due to the release of datasets comprising both topic segmentation and topic identity information \cite{arnold-etal-2019-sector}. Among the most successful systems in this category, S-LSTM \cite{Barrow2020} augmented the hierarchical LSTM with a system to pool sentence embeddings from extracted segments and use the pooled segment representation as input for a topic classification system. Similarly, Transformer$^2_{BERT}$ \cite{Lo2021TransformerOP} used a hierarchical transformer where each contextualized sentence representation is used as input to separate topic segmentation and topic classification classifiers. In all of these cases, the addition of topic class information has been shown to improve results, sometimes quite dramatically. There could be many reasons for this,
but the main rationale is that the shared representation layers in the networks are forced to learn a representation that is similar for all of the sentences sharing a topic class, therefore forcing the model not to focus solely on local cues which often lead to massive overfitting. As a result, adding topic classification in a multi-task setting has been shown to improve the generalizability capacity of topic segmentation models \cite{Lo2021TransformerOP}.

To achieve a similar goal, other works have directly added a secondary loss to segmentation systems, which penalize sentence embeddings belonging to the same topic segment that is too far in the embedding space \cite{xing-etal-2020-improving,yu-etal-2023-improving-long}. Also in this case the use of multi-task learning significantly improved segmentation results.

Another promising research direction is the one of directly injecting the notion of coherence into topic segmentation systems. Coherence modeling 
%itself is a well-understood area of research that has several practical applications and it 
relates quite closely to linear text segmentation in that areas of low coherence in a document often coincide with topic boundaries. Following this reasoning, CATS \cite{Glavas2020} employs a hierarchical transformer built on top of word embeddings and adds a secondary loss in the form of a binary classification where a coherence classification head is tasked with discriminating real text snippets from corrupted ones (i.e.\ text snippets where the sentences have been randomly shuffled). Similarly, Longformer + TSSP + CSSL \cite{yu-etal-2023-improving-long}, the current state-of-the-art in written text segmentation, uses a Longformer as a token-level classifier and adds an auxiliary loss term where a corrupted document having sentences shuffled according to a certain probability is tagged with a series of labels describing whether consecutive sentences are shuffled or not. Both techniques proved to improve results significantly.

Finally, a relative stand-alone recent attempt to combine topic modelling and topic segmentation exists in the form of Tipster \cite{Tipster}, a model that combines neural topic modelling and neural topic segmentation by injecting information from the neural topic model into BERT sentence representations and having them as input for a classic recurrent neural network classifier for segmentation. This is an under-explored area of research that might open interesting future directions.

\section{Datasets}
Many datasets for topic segmentation have been released, but very few have been widely adopted. In this paragraph, we focus on domains that are arguably the most represented in the literature and we divide them in two distinct macro-domains: namely, written text and dialogue. We mostly discuss English datasets, but we will mention in the open challenges the lack of multilingual resources.

\begin{table*}[ht]
    \centering
    \small
    \begin{tabular}{lllllll}
        Name		&Domain		&Language	&\#Documents	&\#Segments per Document	&	\#Sentence per Segment\\
        \hline
		\multicolumn{6}{c}{Written Text}\\
        \hline
        choi		&Random	&English&	920&	9.98 &	7.4 \\
        en\_city		&Wikipedia&	English	&	19500&	8.3	&	56.7\\
        en\_disease 	&Wikipedia&	English	&	3600&	7.5	&	58.5\\
        de\_city 	&Wikipedia&	German	&	12500&	7.6	&	39.9\\
        de\_disease	&Wikipedia&	German	&	2300&	7.2&		45.7\\
        wiki-727k	&Wikipedia&	English	&	727,746& 3.48& 	13.6\\
        %wiki-92k	&Wikipedia&	Multilingual&	90000&	4.41&		4.86\\
        \hline
        		\multicolumn{6}{c}{Dialogue}\\
        \hline
        ICSI		&Meetings&	English	&	25&	4.2	&	188\\
        QMSUM		&Meetings&	English	&	232&	5.54	&	96.93\\
        %MUG		&Meetings&	Chinese	&	360&	9.81	&	101.54\\
        % DialSeg-717	&Conversation&	English	&	711&	-	&	-\\
        % Doc2Dial	&Conversation&	Chinese	&	505&	-	&	-\\
        SuperDialSeg	&Conversation&	English	&	9468&	4.20	&	3.09\\
        TDT		&Media	&	English	&	600*&	88.75*	&	-\\
        Non-NewsSBBC	&Media	&	English	&	54&	7.27	&	72.04\\
        % Radio-NewsSBBC	&Media	&	English	&	48&	11.69&		28.93\\
    \end{tabular}
    \caption{Statistics of some of the datasets discussed. 
    %From the domains, Random refers to randomly concatenated snippets of the Brown Corpus, Wikipedia refers to Wikipedia articles, Meetings, Conversations and Media are transcripts from multi-party meetings, one-to-one conversations and multimedia content respectively. 
    * denotes that the TDT corpus is measured in hours, rather than "number of".}
    \label{tab:stats}
\end{table*}
\vspace{-0.5em}
\subsection{Written Text Datasets}
Written text datasets have been variously proposed over the years, but few have been widely adopted.

Choi was among the first datasets being proposed \cite{choi2000} and it consists of a synthetic dataset created by randomly concatenating sections from different parts of the Brown Corpus. This dataset, however, 
is too simple, which is evident from the fact that an early supervised system like Cross-Segment BERT in table \ref{tab:textresults} was able to get an error already very close to 0. More recently, \citet{Koshorek2018} proposed wiki-727k, a dataset comprising 757,000 Wikipedia articles to overcome the limitations of previous datasets (especially their lack of connection with real use case scenarios) and to provide a dataset big enough to train large supervised models 
like neural networks. This dataset, however, is not widely used as its size makes it expensive to train a full system on it. Most works in topic segmentation, then, currently use en\_city and en\_disease, two English datasets in the Wikisection collection \cite{arnold-etal-2019-sector}, which includes four datasets divided into two categories (articles about cities and articles about diseases) and two languages (English and German); the two datasets are much smaller than wiki-727k and much more focused in terms of domain, where the en\_disease dataset is both the smaller and the more specialized dataset among the two, at it includes a variety of rare medical terms. In general, datasets scraped from Wikipedia have the advantage of not needing any manual annotation, as the headings in the articles are used as topic-shifting markers. At the same time, they present specific challenges as they are composed of portions of texts often written by multiple authors, for which segmentation models might end up recognizing changes in writing style rather than in topics.
\vspace{-0.5em}

\subsection{Dialogue Datasets}
Another active area of research is that of Dialogue Topic Segmentation (DTS), usually in the form of transcripts from multi-party meetings, conversations, podcasts or news shows \cite{Purver2011}.

Initially, datasets for DTS mostly came from the meetings and news shows domains. Early examples of such datasets are the ICSI dataset \cite{icsi}, which includes 70 hours of audio and annotated transcripts from academic meetings, and the TDT corpus \cite{TDT} 
% which was released as part of a series of annual challenges for media engineering and 
including several hundreds of audio and annotated transcripts from American TV news shows. Datasets including transcripts from TV and podcast shows have since been extremely rare and even more rarely datasets were made publicly available mostly due to copyright limitations related to this specific content; TDT itself is available %just after a fee, 
only on paying a fee,
while it is now considered to be too \emph{easy}, as exemplified by the results in table \ref{tab:icsiresults}. Some recent attempts of proposing more challenging, openly available datasets in this domain exist \cite{ICMRGhinassi}, but they are limited in scope and size.
%and much more is needed to foster research in topic segmentation of podcasts and TV shows in general.
QMSUM \cite{zhang-etal-2022-neural} was also recently proposed to collect together different meeting datasets and it includes summary annotation, even though it is considerably smaller than written text-based datasets. 
% On the other side, other datasets including meeting transcripts have been proposed during the years \cite{zhong-etal-2021-qmsum}, even though they are normally much smaller than their written text-based counterparts and the ICSI dataset is still the most widely used in the field.

Finally, one-to-one spoken conversation datasets have been recently proposed. Among these, 
% DialSeg717 overcame the need for expensive manual annotations by randomly concatenating turns from separate conversations about different topics \cite{Xu_Zhao_Zhang_2021}. 
TIAGE was the first manually annotated dataset for one-to-one dialogue, drawing from another existing dataset for NLG \cite{xie-etal-2021-tiage-benchmark}.

Very recently, SuperDialseg was proposed as a large dataset for one-to-one DTS comprising more than 9000 dialogues which were automatically annotated via the use of dialogues that were grounded on the use of written documents in which the separation of topics is known \cite{jiang-etal-2023-superdialseg}. A large meeting dataset was also recently proposed, even though smaller than SuperDialseg, but including annotations for a variety of other tasks \cite{MUGdataset}. These are indeed very promising developments that promise to close the gap between written text segmentation and DTS. Still, more needs to be done in domains such as transcripts from podcasts and TV shows, where comparable resources do not exist. Given the fact that datasets big enough are extremely recent, supervised systems for dialogue segmentation are also rare, even though they have been shown to outperform the alternatives, if enough data are available \cite{jiang-etal-2023-superdialseg}. Table \ref{tab:icsiresults} shows how results on dialogue datasets are similar to the ones obtained on written text datasets by comparable methods; the major challenge in this context, then, is that of having enough data to train supervised systems.

Table \ref{tab:stats} shows statistics from some of the most relevant datasets discussed so far.

\section{Metrics}
\vspace{-0.5em}
Even though traditional classification metrics like F1 and accuracy have been used and continue to be used in the field, specific evaluation metrics for topic segmentations have been variously suggested during the years as traditional classification metrics over-penalize near misses (i.e.\ a topic boundary placed close to a real one), while evidence suggests human annotators tend to disagree where exactly to place topic boundaries \cite{Purver2011}.  

Evaluation in topic segmentation, however, is not a solved problem and specific metrics proposed for evaluating segmentation systems face a number of problems, mostly related to different types of errors (not including a topic boundary, including additional topic boundaries, or placing an existing topic boundary in an incorrect place) and how to quantify and to balance them.

Broadly speaking, segmentation metrics can be categorised into three groups: %, related to their functioning: 
window-based, boundary similarity-based and embedding-based metrics. 

\begin{figure}[!th]
    \centering
    \includegraphics[width=20em, height=20em]{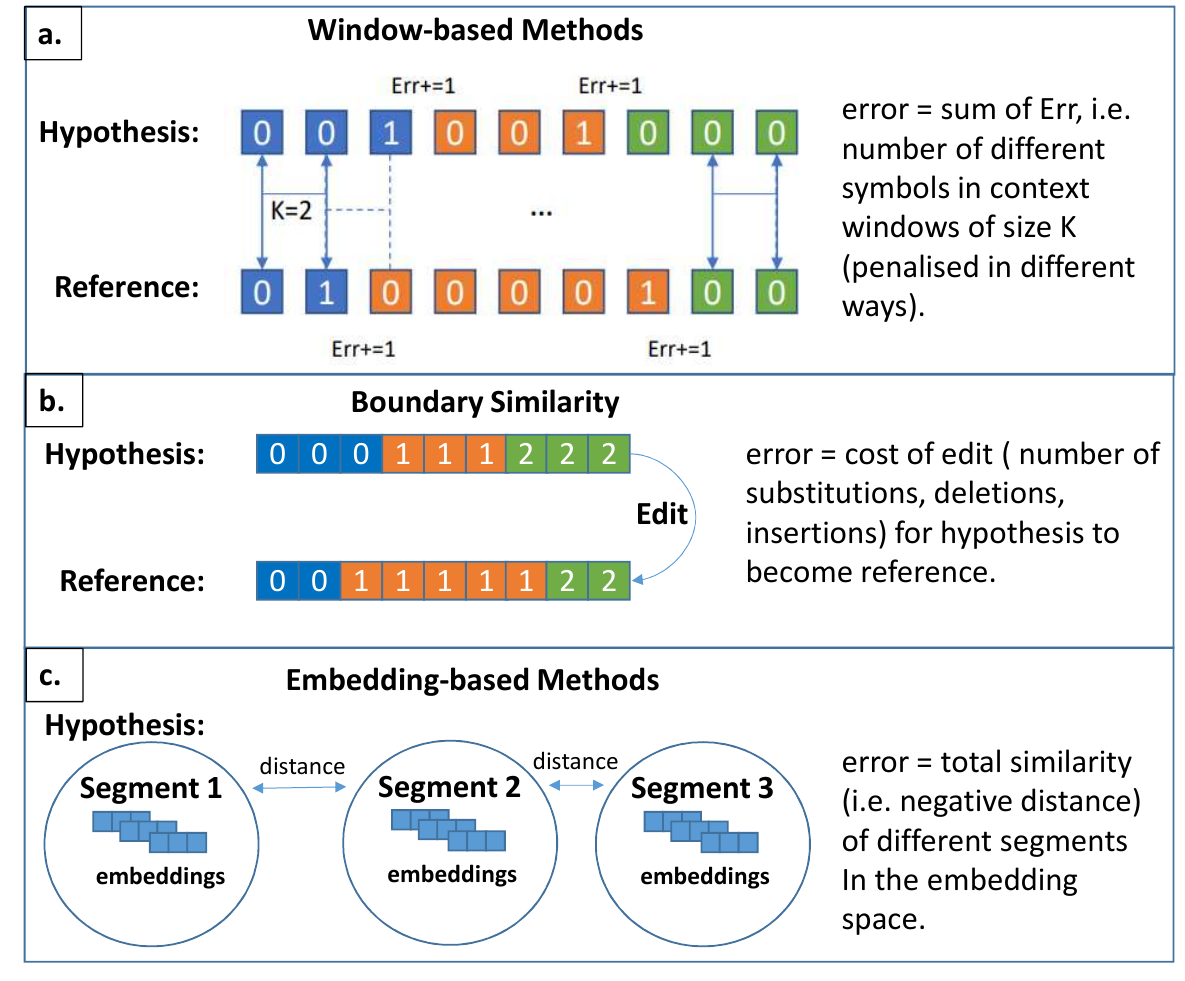}
    \caption{Segmentation metrics comparison. Figure from \citet{ghinassi-etal-2024-cohesion}}.
    \label{fig:metrics-comparison}
\end{figure}

Window-based metrics, exemplified by $P_k$ \cite{Beeferman1999} and WindowDiff \cite{WindowDiff}, employ a sliding window approach, % to assess segment boundaries by 
comparing reference and hypothesis boundaries in the window. 
% Specifically, the $P_k$ metric %works by simply assessing whether the sentences or words at the boundaries 
% measures the likelihood that the hypothesis and reference segmentation agree on whether the start and end of any given window belong to the same or different topic segments. 
%match both in the hypothesis and reference segmentation (i.e.\ they belong to same or different topic segments).
% One of the key weaknesses of these first window-based approaches is that they do not distinguish between false positives and false negatives, while they have been shown to favor false negatives and to be less reliable in certain edge cases \cite{Georgescul2006a}.
To overcome certain limitations of early window-based approaches WinPR \cite{scaiano-inkpen-2012-getting} was proposed to extend the common F1, precision, and recall measures to include a tolerance window.

Drawing on this, Boundary Similarity \cite{fournier-2013-evaluating}, proposed more recently to overcome some of the problems with window-based metrics, works by representing the input sequence using the identity of the topic segment each element in the sequence belongs to. Given such a representation for both the hypothesized and reference segmentation, edit distance is used to quantify the error. 

Finally, reference-free embeddings use notions of embedding similarities to measure similarity within (and/or difference between) hypothesized topic segments, but they still lag behind reference-based metrics in assigning best scores to systems that more closely reflect human-annotated topic boundaries, while also being highly dependent on the quality of the embeddings \cite{SegReFree, ghinassi-etal-2024-cohesion}.

\begin{figure}[h]
    \centering
    \includegraphics[width=25em, height=13em, trim={6cm 0cm 0cm 4cm}]{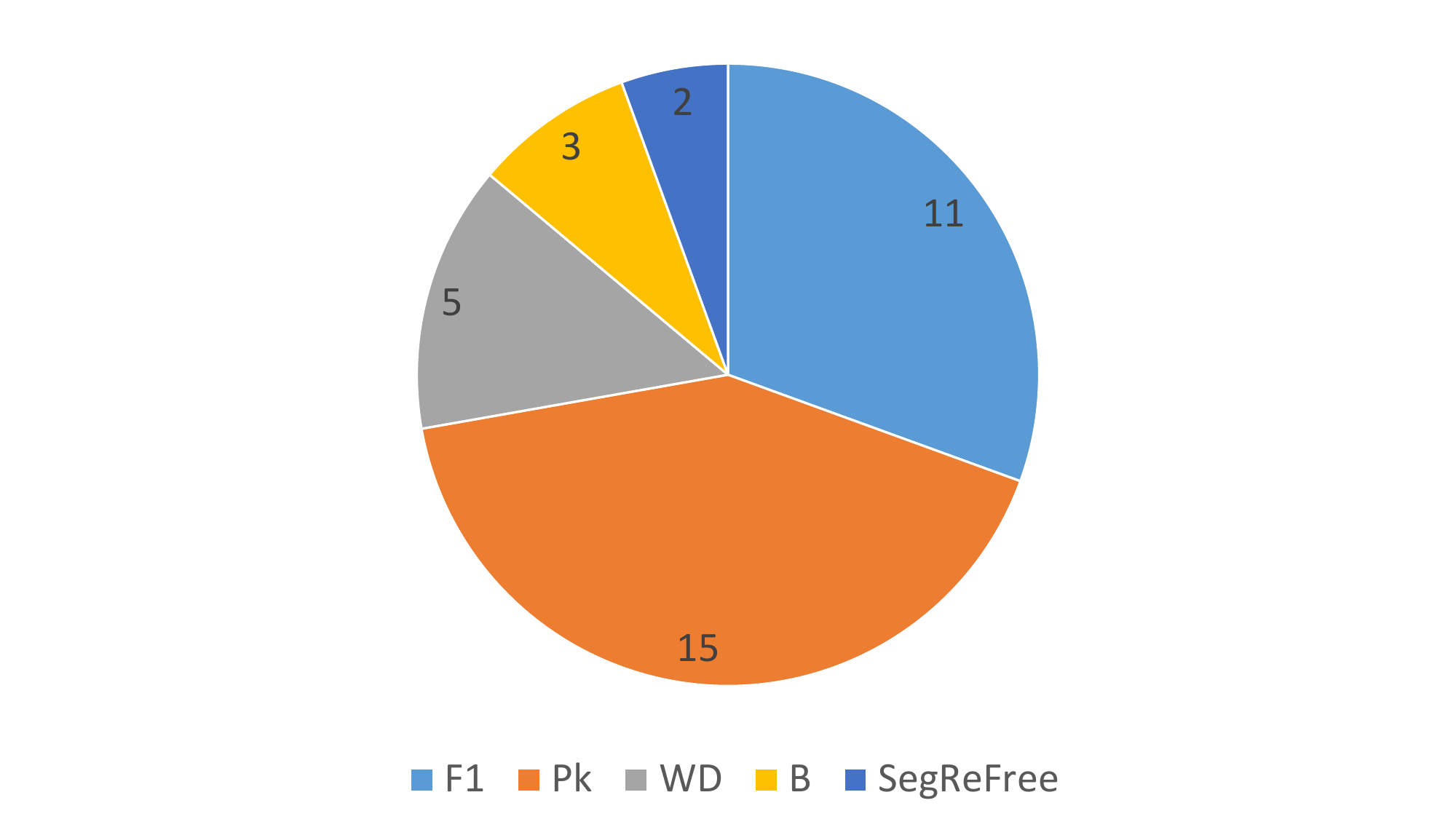}
    \caption{Number of occurrences of F1, Pk, Window Difference (WD), Boundary Similarity (B), and SegReFree in cited works published after 2020.}
    \label{fig:metrics-occurrences}
\end{figure}

\begin{figure}[h]
    \centering
    \includegraphics[width=18em, height=8em, trim={2cm 4cm 0cm 4cm}]{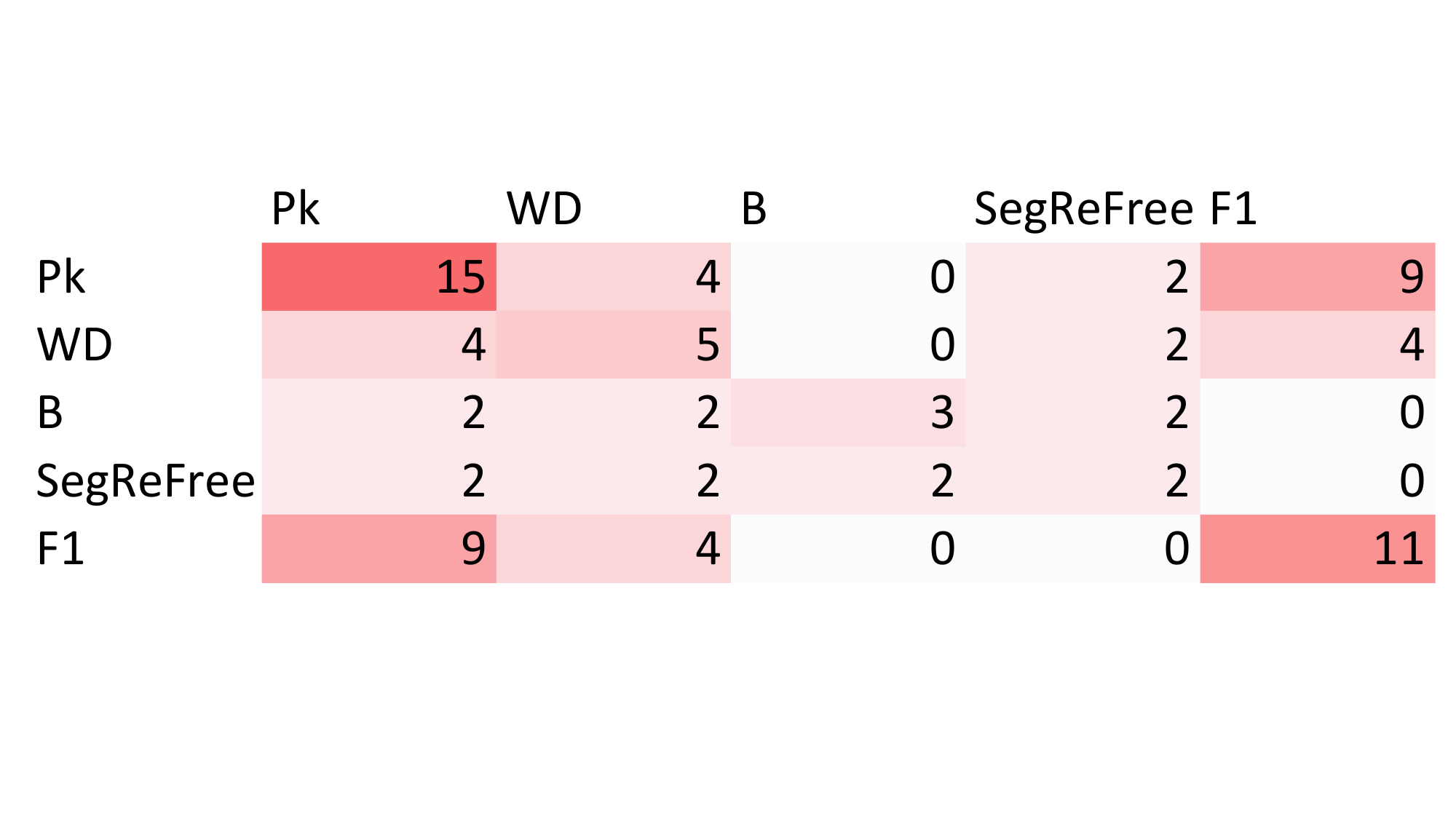}
    \caption{Overlaps between F1, Pk, Window Difference (WD), Boundary Similarity (B), and SegReFree in cited works published after 2020.}
    \label{fig:metrics-overlaps}
\end{figure}

Figure \ref{fig:metrics-comparison} summarises the three different methods just described. $P_k$, WindowDiff, and F1 are the most used metrics in the field. $P_k$ and WindowDiff, however, have been shown to have specific flaws related to penalizing certain types of errors more than others and to behave inconsistently in certain edge cases \cite{Georgescul2006a}. Alternatives like Boundary Similarity, which was proposed to overcome some of the limitations, are not as popular with few works using it and most literature preferring P$_k$, notwithstanding its limitations \cite{Ghinassi2023}. This is evident in figure \ref{fig:metrics-occurrences} showing how popular different metrics are in the literature by the occurrences of different metrics as used in a sample of recent works (i.e.\ published after 2020) we cited. Furthermore, figure \ref{fig:metrics-overlaps} shows that P$_k$ tends to co-occur with WindowDiff and with F1 in recent literature, while it never appeared together with Boundary Similarity. In our systems comparison, We also used P$_k$, but we suggest that future research look into complementing or substituting this metric with more modern ones like Boundary Similarity to overcome well-known evaluation problems with P$_k$ \cite{Georgescul2006a, Ghinassi2023}.

\begin{table*}[ht]
    \centering
    \small
    \adjustbox{max width=\textwidth}{
    \begin{tabular}{llllllll}
        \hline
         Kind  & Basic Unit & System & Choi & en\_city & en\_disease & wiki-727k \\
        \hline
        \multicolumn{6}{c}{Unsupervised Systems} \\
        \hline
        Count-based &  sentence& TextTiling \cite{choi2000}  & 44 & - & - & - \\
        Count-based &  sentence& C99 \cite{choi2000}  & 12 & 36.8 & 37.4 & - \\
        Count-based &  word& U00 \cite{Utiyama2001}  & 9 & - & - & - \\
        Topic Modelling &  word& MM-DP \cite{Misra2011}  & 2.3 & - & - & - \\
        Topic Modelling &  sentence& TopicTiling \cite{Riedl2012}  & 0.95 & 30.5 & 43.4 & - \\
        Embedding-based & word& GraphSeg \cite{Glavas2016} & 7.2 & - & - & - \\
        %\cite{}
        %LLM-based & word & ChatGPT \cite{yu-etal-2023-improving-long} & & & & \\
        \hline
        \multicolumn{6}{c}{Supervised Systems} \\
        \hline
        Single-task & word& TextSeg \cite{Koshorek2018} & - & 24.3 &  19.3 & 22.13 \\
        Single-task & sentence& Cross-segment BERT \cite{Lukasik2020} & \textbf{0.04} & 15.4 &  33.9 & - \\
        % Single-task & word& Seq-ELECTRA-Base \cite{SlidingWindowBERT} & - & - & - & 15.83 \\
        %Single-task & word&Naive LongT5-Base-DS \cite{Inan2022StructuredSU} & - & 8.2 & 33.5 & 15.4 \\
        % Single-task & sentence&Dot-BiLSTM \cite{Ghinassi2023} & - & 8.7 & 20.7 & - \\
        % Multi-task &  word&SEC>T+bloom \cite{arnold-etal-2019-sector}  & - & 14.4 & 26.8 & - \\
        Multi-task &  sentence&Transformer$^2_{BERT}$ \cite{Lo2021TransformerOP}  & - & 9.1 & 18.8 & - \\
        Multi-task &  sentence&Tipster \cite{Tipster}  & - & 8.3 & \textbf{14.2} & - \\
        Multi-task &  word&Longformer + TSSP + CSSL \cite{yu-etal-2023-improving-long}  & - & \textbf{7.4} & 15.4 & \textbf{13.89} \\
    \end{tabular}}
    \caption{Results of various systems described on 4 benchmarks for written text linear text segmentation. Results are reported from the works cited in the table. All results are expressed in P$_k$ metric, the lower the better.}
    \label{tab:textresults}
\end{table*}

\section{Systems Comparison}

Having described unsupervised and supervised approaches for linear text segmentation proposed during the years, table \ref{tab:textresults} and table \ref{tab:icsiresults} present a comparison of performance for different categories described above on some of the benchmarks described in more details in the next section.

In choosing the systems to be compared we have prioritised the inclusion of at least one example for the different kinds of systems described in the previous section (the Kind column in the table) and, where possible, for each system kind we have included at least one example using sentences as basic units and one using words. A final consideration was that of reporting what in our knowledge were the current state-of-the-art for the given benchmarks. 
The choice of systems for which to report results is also limited by the fact that metrics and datasets used vary greatly in existing literature, making the comparison harder.

At first glance, in fact, it can be observed how sparse the tables are: this is due to the long period considered which implies several changes of popular benchmarks over the years, but it also reflects a wider problem in the field for which benchmarks are not consistently used, especially when dealing with domains such as meetings and multimedia content. On another side, it can be seen how supervised models in table \ref{tab:textresults} largely outperform unsupervised systems. Specifically, models based on Longformer which can be trained at the word level as the one by \cite{yu-etal-2023-improving-long} show best performance on most benchmarks. As mentioned, improvements from using multi-task settings seem consistent as most such systems outperform the alternatives, and among those Tipster \cite{Tipster} seems particularly promising. The reason behind such improvements is mostly related to the well-understood problem of supervised systems in topic segmentation, which tend to overfit on local cues and topic shift markers which are by their nature domain-dependent \cite[e.g.\ thanking a correspondent at the end of a news story in news shows,][]{GhinassiPeerJ}. As such, supervised models fail to generalize in many cases. This is even more true in domains in which scarce data is available, which is a common problem to all supervised models but seems to affect even more severely topic segmentation systems \cite{jiang-etal-2023-superdialseg}. Multi-task learning, then, provides a way to direct the model away from focusing on domain-dependent local cues and to focus on properties shared by all sentences belonging to the same topic segments, as it is the case when we combine topic classification and linear text segmentation.

Given the highlighted problem of generalizability, unsupervised systems are still relevant, as the comparatively good performance of BayesSeg on the small ICSI dataset in table \ref{tab:icsiresults} demonstrates. The novel research on the use of LLMs, then, seems particularly relevant as the same table clearly shows how ChatGPT largely outperforms other unsupervised models on the Superdialseg dataset.

\begin{table*}[ht]
    \centering
    \small
    \adjustbox{max width=\textwidth}{
    \begin{tabular}{llllll}
        \hline
         Kind  & Basic Unit & System & ICSI & TDT & SuperDialseg \\
        \hline
        \multicolumn{6}{c}{Unsupervised Systems} \\
        \hline
        Count-based &  sentence& TextTiling \cite{FacebookArticle}  & 38.2 & - & 44.1\\
        %Count-based &  sentence& C99 \cite{Galley2003}  & 33.79 & 9.37 & -\\
        Count-based &  word &  U00 \cite{Galley2003}  & 31.99 & \textbf{4.70} & -\\
        Count-based &  word & BayesSeg \cite{Barzilay2008}  & \textbf{25.8} & - & 43.3\\
        Topic Modelling &  word& HierBayes \cite{Purver2006}  & 28.4 & - & -\\
        Embedding-based & sentence& TextTiling+BERT \cite{FacebookArticle,jiang-etal-2023-superdialseg} & 33.6 & - & 49.9\\
        LLM-based & word & ChatGPT \cite{jiang-etal-2023-superdialseg} & - & - & 31.8\\
        \hline
        \multicolumn{6}{c}{Supervised System} \\
        \hline
        Single-task & word & TextSeg \cite{jiang-etal-2023-superdialseg} & - & - & \textbf{19.9}\\
    \end{tabular}}
    \caption{Unsupervised and supervised systems on benchmarks for dialogue text segmentation. Results are reported from the works cited in the table. All results are expressed in P$_k$ metric, the lower the better.}
    \label{tab:icsiresults}
\end{table*}
\vspace{-0.5em}

\section{Conclusions: Open Challenges and Future Opportunities}
\vspace{-0.5em}
The above discussion 
%of the various methods and resources available for linear text segmentation 
has shown how one of the major challenge in the field is the availability and the adoption of datasets (especially related to DTS). When enough data are available supervised systems can be trained for both written text topic segmentation and DTS generally showing improvements over unsupervised methods. At the same time, the %vast amount
large number
of empty spots in our system comparison tables shows that no single dataset has ever been established as a widely recognized benchmark in the field. Such empty spots are also partly explained by the variety of different metrics for segmentation evaluation, as the lack of a single, widely recognised standard metric means that different works often use different metrics. 
%Many methods explored by the vast literature in the field are either not comparable because no single metric is considered to be standard. 
Moreover, reported performance often does not reflect performance in real-world use cases, because of flaws of existing metrics like P$_k$ \cite{Georgescul2006a}. Future research should, in certain cases like podcast shows segmentation, propose new resources, but mostly it should establish which existing datasets and metrics are best suited to be used as benchmarks and evaluation metrics so that the numerous and rapid advances in this fast-evolving field can be compared in a fair and widely accepted setting.

%Apart from problems with available resources, the available methods for topic segmentation always assume a gold standard having high human annotators agreement, but this is often not the case \cite{Purver2011}. This evidence also stems from the definition of what constitutes a topic: if in domains like news shows identifying a topic can be fairly simple, there are many contexts such as multi-party dialogue in which linearly delimiting topics is more challenging \cite{Purver2011}. Indeed also when segmenting written text like Wikipedia articles a decision needs to be made about which level heading constitutes a big enough topic shift to be considered a topic boundary \cite{Koshorek2018}. Past research has variously tried to move away from linear text segmentation towards a more hierarchical type of segmentation, where segments can be individuated at different topical granularity \cite{HierarchicalTopicSeg,du-etal-2013-topic}; more recent end-to-end systems fall behind under this respect, but promising recent works that combined topic segmentation and topic modelling could draw from the vast literature in hierarchical topic models to aid future research in this sense. Given their outstanding general knowledge, modern LLMs might be particularly suited to combine different tasks in a multi-task and/or zero-shot framework, where initial steps have been already made \cite{ChatGPTTopSeg}.

Apart from resource limitations, methods for topic segmentation often assume a high level of agreement among human annotators, which isn't always the case \cite{Purver2011}. Identifying topics can be straightforward in domains like news shows but more challenging in contexts such as multi-party dialogue. Even when segmenting articles from Wikipedia, decisions must be made about what constitutes a significant enough topic shift \cite{Koshorek2018}. Previous research has explored hierarchical segmentation approaches, moving away from linear text segmentation \cite{HierarchicalTopicSeg,du-etal-2013-topic}. Recent end-to-end systems have lagged in this aspect, but the cited work combining topic segmentation and topic modelling \cite{Tipster} is a promising step forward to exploit knowledge about the topic structure rather than just local cues and coherence. Modern LLMs might be particularly suited to combine different tasks in a multi-task and/or zero-shot framework, as initially explored by \citet{ChatGPTTopSeg}.

% The idea of multi-task and transfer learning was also discussed in the context of solving the generalization problem highlighted by existing literature \cite{GhinassiPeerJ}. In this sense, the role that unsupervised models can play in the field is likely not marginal and, especially, advances in LLMs can be crucial in advancing this strand of research, as initial experiments exploiting the zero-shot capabilities of such model have shown some success \cite{ChatGPTTopSeg}.   

%Finally, our discussion focused mostly on English resources. Recently more linguistically diverse resources have been proposed, especially for Mandarin \cite{MUGdataset}, while we mentioned the presence of two popular German datasets \cite{arnold-etal-2019-sector}. Not many other examples of topic segmentation for other languages exist, however, with a notable exception being the multi-language dataset proposed by \cite{multilingual_wikitopic_dataset}, which has not been widely adopted. Expanding the field to multilingual settings, then, is crucial in line with much recent NLP research to democratize and expand the scope of the field.

Our discussion primarily focused on English resources. Recently, more diverse linguistic resources have been suggested, especially for Mandarin \cite{MUGdataset}, with two German datasets also noted \cite{arnold-etal-2019-sector}. Few examples of datasets for other languages exist, except for the multilanguage dataset proposed by \cite{multilingual_wikitopic_dataset}, which remains underutilized. Multilinguality is crucial to democratize and broaden the scope of NLP research.

To summarise, in this work we have traced the various existing trends in literature for linear text segmentation within NLP and we have identified the following main challenges:
% \begin{enumerate}
%     \item Lack of publicly available datasets: this problem affects mostly DTS (specifically the media domain) and it is crucial as recent supervised systems greatly outperform unsupervised ones. As a subset of this problem, we have also mentioned the need for standard benchmarks for the task to better track the advances in the field.
%     \item Pitfalls in existing metrics: we have discussed how the most popular metric, P$_k$ has a number of well-documented shortcomings. Even though newer metrics like Boundary Similarity have been proposed, P$_k$ is the most used also in recent works and there is a lot of scope to adopt or devise better metrics.
%     \item Low generalizability: we have also discussed how the field has individuated generalizability as a key problem for the task, as many well-performing supervised systems might just be overfitting on specific cue phrases.   
% \end{enumerate}

\textbf{Lack of publicly available datasets:} this problem affects mostly DTS (specifically the media domain) and it is crucial as recent supervised systems greatly outperform unsupervised ones. As a subset of this problem, we have also mentioned the need for standard benchmarks for the task to better track the advances in the field.

\textbf{Pitfalls in existing metrics:} the most popular metric, P$_k$ has a number of well-documented shortcomings. Even though newer metrics like Boundary Similarity have been proposed, P$_k$ is the most used even in recent works. 
%and there is a lot of scope to adopt or devise better metrics.

\textbf{Low generalizability} we have also discussed how the field has individuated generalizability as a key problem for the task, as many well-performing supervised systems might just be overfitting on specific cue phrases. 

We suggest the following future directions as open opportunities for researchers in the field:
% \begin{enumerate}
%     \item Use of LLMs: the rise of LLMs has already reshaped many areas in NLP, and there is similar scope in this context, especially given the problems of generalizability and the lack of resources which affect the field.
%     \item Advances in Multi-task learning: we have seen how the multi-task learning paradigm has been widely used recently in the field. Still, the majority of works tend to combine topic segmentation with topic classification, while we highlighted how the combination of modern segmentation systems with topic modelling ones is a research direction worth developing, as it has deep roots in the field and it can open further research in hierachical segmentation, which in turn is useful for overcoming the problem of arbitrary definition of topic boundaries.
%     \item Advances in evaluation resources and metrics: finally we stress the importance of having a stable and broad evaluation framework for the task. Under this respect advances in metrics are particularly welcome as they might also help deepen our understanding of a task which often sees human annotators disagreeing, while multi-lingual datasets might help widening the reach of the available technology to less-resourced languages.
% \end{enumerate}

\textbf{Use of LLMs:} the rise of LLMs has already reshaped many areas in NLP, and there is similar scope in this context, especially given the problems of generalizability and the lack of resources which affect the field.

\textbf{Advances in Multi-task learning:} 
%we have seen how the multi-task learning paradigm has been widely used recently in the field. Still, the majority of works tend to combine topic segmentation with topic classification, while we highlighted how 
we highlight the combination of modern segmentation systems with topic modelling ones as a research direction worth developing, having deep roots in the field and narrowing the gap with hierachical segmentation, which is useful for overcoming the problem of arbitrary definition of topic granularity.

\textbf{Advances in evaluation resources and metrics:} we stress the importance of having a stable evaluation framework for the task. Advances in metrics are useful to deepen our understanding of a task having low human annotators agreement. Multi-lingual datasets, instead, can widen the reach of the available technology to less-resourced languages.

% Recent years have indeed seen a number of promising steps in the field and this work highlights various directions and starting points to tackle such challenges.

\section{Limitations}
Our work aimed to fill noticeable gaps in literature on topic segmentation. As previous surveys on the topic are all outdated or limited in scope, the current survey does not cover some of the many advances in the field explored in recent years.
Among them, in our work we did not cover:
\begin{enumerate}
    \item Multi-modality.
    \item Topic Segmentation in nicher domains, like educational and legal text and multimedia.
    \item Graph based methods for Topic Segmentation.
\end{enumerate}

Another limitation of our work involves the definition of the classes for topic segmentation. In presenting an overview of available metrics, in fact, we have picked popular metrics for topic segmentation, but we have left out less used metrics that have been proposed and that might not fall neatly in the three-fold division of available methods that we have proposed.

Finally, we have mentioned the existing limitations of topic segmentation for languages other than English. Our work mostly deals with English resources, even though it mentions at least some literature dealing with other languages. This limitation is partly due to limitations within the field, which we have mentioned in our conclusions, but future work might integrate more research in this direction.

\section*{Acknowledgements}

Purver was funded by the UK EPSRC (project ARCIDUCA, EP/W001632/1), Responsible AI UK (keystone project AdSoLve) and the Slovenian Research Agency (research core funding P2-0103 and project EMMA L2-50070).

% \bibliographystyle{named-short}
\bibliography{custom}

\end{document}